# A Multi-Source Heterogeneous Knowledge Injected Prompt Learning Method for Legal Charge Prediction

Jingyun Sun[1], Chi Wei[2], Yang Li[1,*]

College of Computer and Control Engineering, Northeast Forestry University, Harbin, China.
College of Computer Science and Technology, Harbin Engineering University, Harbin, China.

Corresponding author:   yli@nefu.edu.cn

## Abstract

Legal charge prediction, an essential task in legal AI, seeks to assign accurate charge labels to case descriptions, attracting significant recent interest. Existing methods primarily employ diverse neural network structures for modeling case descriptions directly, failing to effectively leverage multi-source external knowledge. We propose a prompt learning framework-based method that simultaneously leverages multi-source heterogeneous external knowledge from a legal knowledge base, a conversational LLM, and related legal articles. Specifically, we match knowledge snippets in case descriptions via the legal knowledge base and encapsulate them into the input through a hard prompt template. Additionally, we retrieve legal articles related to a given case description through contrastive learning, and then obtain factual elements within the case description through a conversational LLM. We fuse the embedding vectors of soft prompt tokens with the encoding vector of factual elements to achieve knowledge-enhanced model forward inference. Experimental results show that our method achieved state-of-the-art results on CAIL-2018, the largest legal charge prediction dataset, and our method has lower data dependency. Case studies also demonstrate our method's strong interpretability.

**Keywords**: Legal charge prediction; knowledge-enhanced prompt learning; contrastive learning

# 1. Introduction

Legal charge prediction is a crucial task in legal artificial intelligence, aimed at utilizing advanced technologies such as machine learning, deep learning, and natural language processing to analyze given case descriptions and thereby predict corresponding charge labels. Figure 1 presents an example of the task. The figure provides a case description, from which it can be inferred that the case pertains to a mobile phone theft. Consequently, the "theft" label is selected from the candidate legal charge labels to be assigned to this case description.

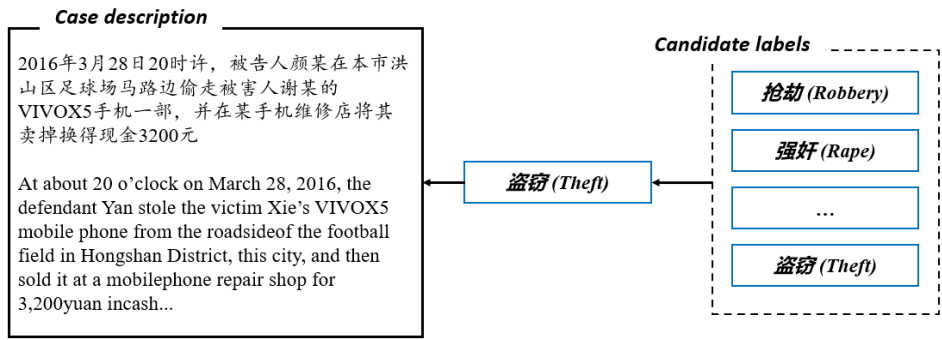

**Fig. 1** An example of the legal charge prediction task

Legal charge prediction not only helps legal professionals handle cases efficiently and accurately, reducing human errors and increasing the consistency and fairness of judgments, but also supports the enhancement of legal education and public legal awareness, by spreading legal knowledge and strengthening society's understanding and compliance with legal regulations.

Legal charge prediction is often regarded as a classification problem, hence researchers typically adopt methods similar to those used for general text classification tasks to address it. For instance, Wang et al. proposed a convolutional neural networks-based approach [1], Yang et al. introduced a method based on bidirectional long short-term memory network [2], and Chen et al. developed a gated reccurent units-based method [3]. However, legal charge prediction differs from general text classification tasks in several ways. Firstly, legal texts, which contain a plethora of legal terminologies and keywords, are distinct from general texts, presenting challenges to universal models in understanding the content. Secondly, legal charge prediction focuses more on the factual information within texts, whereas general text classification tasks are concerned with the topics described by texts. Therefore, many researchers have utilized language models pretrained in the legal domain as the backbone to enhance the model's comprehension of legal texts [4]. Such models are better at capturing the domain-specific terminologies and keywords within legal texts. Nevertheless, the pretrained models they employed only allow for the input of 512 tokens, which is insufficient for modeling legal texts that often exceed this length. Moreover, to capture factual information in legal texts, Sukanya & Priyadarshini et al. proposed a model based on attention mechanism [5], and Wang et al. further introduced a hierarchical attention mechanism to capture factual information at different levels [6]. However, they only utilized the content of legal texts themselves to obtain factual information, without leveraging external structured knowledge.

In contrast, the method we propose not only enhances the model's comprehension of the textual content and task objectives but also fully leverages heterogeneous external legal knowledge from multiple sources. Firstly, we employ the newly introduced pre-trained language model, Lawformer, which is trained on a large-scale legal corpus and can accommodate text inputs exceeding 4,000 tokens, as our inference model. Lawformer aids in accurately comprehending the semantics of case descriptions and capturing the meanings expressed by legal terms and keywords. Subsequently, we utilize a legal knowledge base to match knowledge snippets from case descriptions, while employing a conversational LLM and relevant articles to extract factual elements from descriptions. This process introduces external components to assist the model in acquiring legal knowledge, thereby further enhancing the model's understanding of case descriptions. Finally, we propose using soft prompt tokens and hard prompt templates to encapsulate heterogeneous legal knowledge from multiple sources. Overall, the method presented in this paper leverages the paradigm of prompt learning to integrate heterogeneous legal knowledge from multiple sources into the model's forward reasoning, thus improving the model's predictive accuracy regarding legal charges.

We conducted extensive experiments on CAIL-2018, the largest legal charge prediction dataset to date [7]. The results demonstrate that our proposed method achieved results surpassing the baselines, with a macro F1 score of 0.84. Moreover, the experiments indicate that our method has the lowest dependency on training data. The performance of other baselines significantly diminishes as the scale of training data decreases, whereas our method still maintains a high F1 score. We also analyzed the contribution of each module within our method through ablation studies. Finally, we validated that our approach possesses strong interpretability, which is crucial for artificial intelligence tasks in the legal domain.

Our primary contributions can be summarized as follows:

1) We propose a legal charge prediction model that integrates multi-source heterogeneous legal knowledge.

2) We introduce a method to encapsulate heterogeneous legal knowledge via the prompt-based learning framework.

3) We propose the use of a conversational LLM and relevant legal articles to extract factual elements from case descriptions.

4) We employ a specialized legal knowledge base to match knowledge snippets from case descriptions.

5) We demonstrate the effectiveness of our method through extensive empirical validation.

## 2. Related work

This section presents the works related to our study. Firstly, the latest progress in legal charge prediction is introduced. Subsequently, the basic concepts and applications of prompt learning are introduced.

### 2.1 Legal charge prediction

Legal charge prediction is a crucial task in the field of legal artificial intelligence, aimed at predicting the legal charges corresponding to given case descriptions. Due to the scarcity of legal resources, individuals lacking legal knowledge often find it challenging to promptly seek legal advice from attorneys or legal professionals. Therefore, the automation of legal charge prediction can, to a certain extent, alleviate the issue of legal resource scarcity. Furthermore, legal charge prediction can also offer decision support for lawyers or judges, thereby enhancing their work efficiency.

Early legal charge predictions primarily relied on rule-based methods or mathematical models [8], [9], [10]. These methods have the advantage of transparent and intuitive reasoning processes, and once the inference rules are triggered, their outcomes are fixed. However, such methods exhibit poor generalization and struggle to effectively address language phenomena such as synonyms and polysemy in case descriptions. With the introduction of the Word2Vec concept by Mikolov [11], subsequent legal charge prediction methods have predominantly been based on semantic embeddings.

These methods embed words in case descriptions into semantic vectors, which are then used as features inputted into a machine learning model [4], [12]. An exemplary contribution is Law2Vec proposed by Chalkidis et al [13]. Law2Vec is a specialized word embedding for the legal domain, trained on 123,066 legal documents containing 492M words. Building upon Law2Vec, Chalkidis et al. introduced a logistic regression-based method [14]. This method is simple yet effective, demonstrating an F1 score over 40% higher than rule-based methods on an EU legislative dataset. Additionally, methods combining semantic embeddings with support vector machines and k-nearest neighbors have been proposed [15].

With the popularity of deep learning technologies, researchers have shifted from merely combining semantic embeddings with shallow machine learning models to integrating them with neural network models [4], [16], [17], [18]. Compared to shallow machine learning models, neural network models exhibit stronger data fitting and feature learning capabilities, leading to superior performance. For instance, Wang et al. proposed a method based on convolutional neural networks [1]. This method leverages the modeling capability of convolutional networks for local key information, enabling the model to identify crucial phrases, terms, jargon, and keywords in case descriptions, thereby enhancing the performance for legal charge prediction. On the other hand, Yang et al. focused on modeling the global semantic correlations within case descriptions and introduced a method based on long short-term memory networks [2]. To reduce the computational complexity of globally modeling semantic relationships in case descriptions, Chen et al. presented a method based on gated recurrent units [3]. This method employs computationally less intensive gated recurrent units to model the global semantic correlations of case descriptions, significantly reducing computational complexity. Additionally, Sukanya & Priyadarshini proposed a model based on attention, which can attend to salient information in different aspects of a case description [5]. Building upon this, Wang et al. introduced a hierarchical attention model capable of attending to salient information at different levels within a case description [6]. This approach achieved state-of-the-art performance in various benchmarks. Despite the generally excellent results achieved by neural network-based models, they are still constrained by the effectiveness of word embeddings and rely heavily on large amounts of high-quality annotated data.

In recent years, an increasing number of methods for predicting legal charges have been based on pre-trained language models [4]. Unlike neural network models that take static semantic embeddings as input, pre-trained language models are pre-trained on large-scale textual data and therefore have better context understanding capabilities [19], [20], [21]. Moreover, pre-trained models can capture relationships and contexts between different words in a case description, thereby better comprehending the textual meaning of legal narratives and contributing to improved accuracy in charge prediction [22]. The core of our method lies in a pre-trained legal language model, ensuring its contextual understanding of legal texts. Diverging from existing methods, we integrate heterogeneous legal knowledge from multiple sources into the reasoning process of the language model, enabling it to acquire a more comprehensive understanding of legal knowledge specific to a given case description. Furthermore, compared to traditional neural network methods, our approach exhibits lower data dependency.

**2.2 prompt learning**

Prompt learning has recently garnered significant attention from researchers due to its ability to stimulate language models to better recall the semantic knowledge learned during pre-training [23], [24]. Unlike the standard downstream task fine-tuning paradigm, the prompt learning paradigm aligns downstream tasks with the pre-training tasks of language models. To this end, methods based on prompt learning should first convert different downstream tasks into a language modeling task [25], [26]. For instance, a traditional classification task is designed to fit the probability distribution $Y = (X; \theta)$. Given a piece of text $x = [This\ pizza\ is\ so\ delicious]$, the model might output the prediction $y = 0 \in \{0,1,2\}$ once $\theta$ is learned. Where $0$ denotes a positive sentiment label, $1$ denotes a negative sentiment label, and $2$ denotes a neutral sentiment label. However, the model aims to fit the function $Y = P(MASK = \mathcal{V}_y | template(X); \theta)$ by converting the task into a language modeling task. Here, $template(x)$ is a new text transformed from the original text by inserting specific prompt words and $\mathcal{V}_y$ is the set of label words. For example, the original text $x = [This\ pizza\ is\ so\ delicious]$ could be transformed into $template(x)$=[This pizza is so delicious. It feels [MASK]], and the model is tasked with predicting the word at the [MASK] position based on $\theta$, thereby inferring the sentiment label. The model might generate words such as "amazing", "bad", or "okay", which can then be mapped to the specific sentiment labels 0, 1, or 2.

The three core components in prompt learning are prompt templates, inference models, and label mappings. A ***prompt template*** is employed to encapsulate a original text into a new format featuring prompts and masks, as exemplified in the work of [27], [28], [29]. They integrate external knowledge at the prompt template stage to maximize the language model's understanding of the task. Our method also incorporates external knowledge during the prompt template stage. However, we propose the integration of multi-source heterogeneous knowledge into the prompt template unlike existing methods, thereby significantly enhancing the model's inference capabilities. ***Inference model*** is a core component of prompt learning, utilized for predicting the tokens at the mask positions based on the encapsulated text. Commonly used inference models include BERT [30], RoBERTa [31], the GPT series [32], and the T5 series [33]. Furthermore, some studies employ language models specialized in specific domains to cater to particular tasks. For instance, Zhu et al. proposed using CliniBERT [34] for prompt learning methods in the medical field, achieving state-of-the-art results. We utilize Lawformer, a pre-trained language model in the legal domain, as the inference model, making our method more suitable for legal charge prediction tasks.

## 3. Task formalization

The task of legal charge prediction aims to accomplish the following process: given a case description $X$ containing $L$ tokens, the model predict a legal charge label $\hat{y}$ based on the content of $X$. This process can be denoted as Equation 1.

$$\hat{y} = Model(X; \theta) \tag{1}$$

Where $\theta$ represents the learnable parameters within the model.

In this study, we additionally utilize a set $\mathcal{A}$ of legal articles, a conversational large language model $LLM(\cdot)$, and a legal knowledge base $\mathcal{K}$ as aids, thereby resulting in the Equation 2.

$$\hat{y} = Model(X,\ \mathcal{A},\ LLM(\cdot),\ \mathcal{K};\ \theta) \tag{2}$$

## 4. Methodology

Figure 2 illustrates our method, comprising four modules. ***The first module***, as depicted in the lower-left corner of the figure, focuses on acquiring and encoding factual elements from the given case description $X$. In this module, case description $X$ search for the most relevant $N$ legal articles via a joint semantic space, subsequently consulting a conversational Large Language Model (LLM) for factual elements in the case description based on these legal articles. The factual elements aquired are encoded into a semantic vector $\vec{u}$ by a BiGRU encoder, and $\vec{u}$ is then injected into the forward computation of the soft prompt tokens to enhance the model's reasoning capabilities for the task. ***The second module*** involves knowledge matching based on a legal knowledge base, wherein knowledge from the case description $X$ is matched with a knowledge base $\mathcal{K}$. The knowledge snippets matched are then concatenated and added to the original input as prompt to further strengthen the model's reasoning. ***The third module*** is the cornerstone of our method consisting of a legal language model, as illustrated in the central part of the figure. The input to this legal language model includes five components: 1) two soft prompt tokens $s_1$ and $s_2$; two manually constructed template texts $T_1$ and $T_2$; 3) the masked tokens $M$; 4) the case description $X$; and 5) the knowledge snipptes $K$. The objective of the legal language model is to predict the tokens at the masked positions. ***The fourth module*** aims to map the predictions of the language model at the masked positions onto a legal charge category $\hat{y}$, serving as the final process of our method.

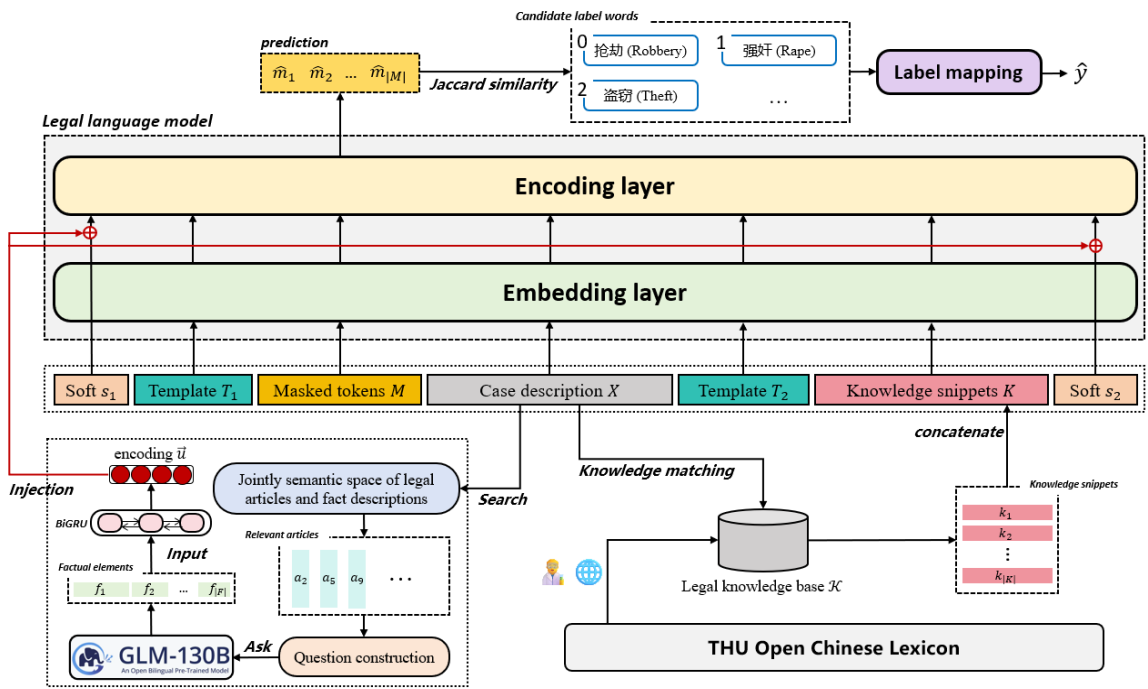

**Fig. 2** Model architecture diagram

In the following sections, we detail each module: Section 4.1 covers the first module, Section 4.2 the second, Section 4.3 the third, and Section 4.4 the fourth.

## 4.1 Factual elements acquisition and encoding

This section introduces the first module of our method, which involves acquiring and encoding factual elements from the given case description. Factual elements in case descriptions are crucial for legal judgments, as they influence the overall understanding of the cases and the final verdicts [35]. These elements typically include the time and place of the event, the individuals involved, and the specific process of the event. Example 1 demonstrates a case description and its contained factual elements.

**Case description:**

柯某某在未经著作权人授权的情况下，采用"火车采集器"网络爬虫软件，从视频网站采集5万余部电影，存储在租用的服务器上。柯某某将存储在服务器的影视作品转载到其个人运营管理的网站上提供给网民免费观看，同时收取广告费，非法获利共计人民币35万余元 (Without the authorization of the copyright owners, Ke employed "Train Collector" web crawling software to collect over 50,000 movies from a video website, storing them on a rented server. Ke then reposted these movies on a website personally operated and managed, offering them for free viewing to netizens. Concurrently, he collected advertising fees, illegally profiting a total of over 350,000 yuan)

**Factual elements:**

- Subject: Ke
- Authorization: Ke acted without the authorization of the copyright holders.
- Tool used: Utilized "Train Collector" web crawling software.
- Storage method: Stored the works on a rented server.
- Profit method: Earned revenue through advertising fees.
- Illicit Gains: Illegally profited a total of over 350,000 yuan.

**Example 1** A case description and its contained factual elements

In our method, the acquisition and encoding of factual elements involve the following four processes. Firstly, we utilize the case descriptions and legal articles from the entire training set to learn a joint semantic space. Then, when a case description $X$ is given, this joint semantic space is employed to find the $N$ legal articles most relevant to it. Subsequently, these $N$ legal articles are used to consult a conversational LLM about the most noteworthy factual elements in the case. Finally, the obtained factual elements are encoded into a semantic vector $\vec{u}$. The following sections will elaborate on these processes.

**4.1.1 Jointly semantic space learning**

We manage to utilize relevant legal articles to extract noteworthy factual elements from case descriptions, as these articles explicitly define which factual elements are pertinent to specific legal charges. For instance, the legal article pertinent to the crime of copyright infringement is: "*Acts such as copying and distributing literary, audio-visual, and computer software works for profit without the permission of the copyright holder, publishing books that are subject to another's exclusive publishing rights without consent, duplicating audio-visual products created by others without their permission, producing and exhibiting art works falsely attributed to another, where the amount of illegal gains is substantial or other serious circumstances are presen*t". From this, we can infer that elements such as whether the intent was for profit, whether copyright permission was obtained, and the amount of illegal gains, are factual aspects worthy of attention.

To match relevant legal articles with case descriptions, we propose the construction of a joint semantic space of both case descriptions and legal articles. We engage in contrastive training of the language model RoBERTa [31] to facilitate its learning of this joint semantic space. Compared to rule-based methods [36], [37], the language model can model deeper semantic connections between case descriptions and legal articles, thereby achieving superior matching outcomes. Furthermore, contrastive training places greater emphasis on the relative relationship between positive and negative samples compared to traditional neural network-based semantic matching methods [38]. Consequently, contrastive training aids RoBERTa in learning more distinct and discriminative features which are crucial in determining the relevance between case descriptions and legal articles. Next, we introduce the specific steps of using RoBERTa to learn the joint semantic space.

*Step 1: Construct positive and negative pairs*

Each case description in CAIL-2018, the largest dataset of legal charge prediction, has been labeled its relevant legal articles. Therefore, we can easily construct contrastive positive and negative pairs from the entire training set automatically. Given the training set containing the pairs of case descriptions and relevant legal articles $\mathcal{D}_{Train} = \{(X_1, \mathcal{R}_1), (X_2, \mathcal{R}_2), (X_3, \mathcal{R}_3), ...\}$, we use Algorithm 1 to automatically construct a set $\mathcal{P}$ of positive and negative pairs. Where, $X_i$ represents the $i_{th}$ case description in the training set, and $\mathcal{R}_i$ represents the legal articles related to it.

---

**Algorithm 1:** Construction of the positive and negative pair set

**Input:** Training set $\mathcal{D}_{Train} = \{(X_1, \mathcal{R}_1), (X_2, \mathcal{R}_2), (X_3, \mathcal{R}_3), ...\}$
**Output:** Set $\mathcal{P}$ of positive and negative pairs

1   **Initialize** an empty set $\mathcal{P} = \{\}$
2   **for** each $(X_i, \mathcal{R}_i)$ in $\mathcal{D}_{Train}$ **do**
3       $C \leftarrow 0$         //Count the number of positive pairs
4       **for** each legal article $r$ in $\mathcal{R}_i$ **do**
5           Add $(X_i, r)$ into $\mathcal{P}$       //$(X_i, r)$ is a positive pair
6           $C \leftarrow C + 1$
7       **end for**
8       Randomly select $C$ number of $\mathcal{R}^{neg}$ from $\complement_{\mathcal{D}_{Train}}(X_i, \mathcal{R}_i)$
9       Select a legal article $r^{neg}$ from each $\mathcal{R}^{neg}$
10      Pair $X_i$ with each legal article $r^{neg}$: $(X_i, r^{neg})$
11      Add $C$ number of $(X_i, r^{neg})$ into $\mathcal{P}$       //$(X_i, r^{neg})$ is a negative pair
12   **end for**
13   **return** $\mathcal{P}$

---

*Step 2: Obtain representations of case descriptions and legal articles*

We have obtained set $\mathcal{P}$ of positive and negative pairs in the first step. Each pair in the set is either a related pair of case description and legal article or is irrelevant. In this step, we employ RoBERTa to acquire semantic vectors for the case description and legal article in each pair. This process is illustrated in Equation (3).

$$\boldsymbol{P} = RoBERTa(\mathcal{P}) \tag{3}$$

Herein, $\boldsymbol{P}$ represents a matrix, where the odd-numbered columns of $\boldsymbol{P}$ denote the semantic vectors of case descriptions, and the even-numbered columns represent the semantic vectors of legal articles. Now that we have obtained the semantic vectors for all samples in positive and negative pairs, we proceed to train RoBERTa using a contrastive loss.

*Step 3: Train the RoBERTa via contrastive loss*

During the training, RoBERTa learns to decrease the semantic distance between samples in positive pairs while increasing the distance between those in negative pairs. This objective is achieved through a contrastive loss function, which quantifies the similarity between semantic vectors in a pair. The calculation of the loss for the $i_{th}$ case description is as shown in Equation (4).

$$\ell_i = -\log \frac{\sum_{c=1}^{C} e^{sim(\vec{p}_i, \vec{p}_c^+)/\tau}}{\sum_{c=1}^{C}(e^{sim(\vec{p}_i, \vec{p}_c^+)/\tau} + e^{sim(\vec{p}_i, \vec{p}_c^-)/\tau})} \tag{4}$$

Where, $\vec{p}_i$ represents the semantic vector of the $i_{th}$ case description, while $\vec{p}_c^+$ and $\vec{p}_c^-$ respectively denote the semantic vectors of the legal articles in the $c_{th}$ positive and negative pairs. Besides, $sim(\vec{p}_1, \vec{p}_2)$ denotes the cosine similarity between vectors $\vec{p}_1$ and $\vec{p}_2$, and $\tau$ is a temperature hyperparameter.

The trained RoBERTa model can encode case descriptions and legal articles into a joint semantic space, where the representation of a case description and its corresponding legal articles exhibit a closer semantic distance within this space.

### 4.1.2 Relevant legal articles searching

Having obtained a joint semantic space for case descriptions and legal articles through prior operations, this section utilizes this joint semantic space to search $N$ legal articles relevant to a given case description. Given the case description $X$ and a set of candidate legal articles $\mathcal{A}$, we first encode them into the joint semantic space using the trained RoBERTa model, obtaining their respective semantic vectors. This process is illustrated in Equation (5).

$$\vec{x}, [\vec{a}_1, \vec{a}_2, \ldots, \vec{a}_{|\mathcal{A}|}] = RoBERTa(X, \mathcal{A}) \tag{5}$$

Where, $\vec{x}$ represents the semantic vector of case description $X$, and $\vec{a}_i$ denotes the semantic vector of the $i_{th}$ legal article in set $\mathcal{A}$.

Subsequently, we compute the relevance between the case description and each candidate legal article through vector inner product, as shown in Equation (6), thereby selecting the $N$ legal articles with the highest relevance.

$$Relevance(\vec{x}; \vec{a}_i) = \vec{x} \cdot \vec{a}_i \tag{6}$$

In the next section, we will utilize the $N$ legal articles searched, along with case description $X$, to consult a conversational large language model. This is done to acquiring noteworthy factual elements within $X$.

### 4.1.3 Conversational large language model consultation

| Question template |
|---|
| Messeges= |
| [{"role": "user", "content": "Factual elements in a case description refer to: specific facts used to describe and prove the circumstances of the case, including basic information such as time, location, characters, and the sequence of events."}, |
| {"role": "assistant", "content": "That's correct. Factual elements in a case description are indeed the specific details used to outline and substantiate the circumstances of a case."}, |
| {"role": "user", "content": "Please analyze the case description in < > based on the legal articles in << >>, and list 5-10 factual elements into [ ]"}] |

Conversational LLMs, with their vast parameter count, possess robust contextual reasoning capabilities and have learned a wealth of generic world knowledge during their pre-training. Furthermore, Conversational LLMs often exhibit strong zero-shot reasoning capabilities, thus enabling their direct use as ready-made tools without the need for additional fine-tuning. Based on these reasons, we use a Conversational LLM to assist us in acquiring factual elements from $X$. We construct the following question template.

We utilize this template to conduct inquiries with the conversational LLM, resulting in a list of factual elements denoted as $F = [f_1, f_2, \ldots, f_{|F|}]$. These factual elements are subsequently be encoded as semantic vectors in Section 4.1.4. When the dataset is in Chinese, we utilize GLM-130B, a Chinese conversational LLM developed by ZhiTu HuaZhang Technology Co., Ltd., in this process. When the dataset is in English, we directly invoke the API of ChatGPT to achieve this process.

**4.1.4 Factual elements encoding**

Given the list of factual elements, $F = [f_1, f_2, ..., f_{|F|}]$, obtained from the previous process, we concatenate them and input the combined sequence into a BiGRU encoder. A BiGRU consists of two GRU layers that process the data in opposite directions: one forward GRU and one backward GRU. The forward GRU processes the sequence from $f_1$ to $f_{|F|}$, and the backward GRU procsses it from $f_{|F|}$ to $f_1$. Each GRU updates its hidden state at each step in the sequence.

Let $\vec{h}_t$ be the hidden state of the forward GRU at time step $t$, and $\overleftarrow{h}_t$ be the hidden state of the backward GRU at time step $t$. They are computed as Equations (7-8).

$$\vec{h}_t = GRU(f_t, \vec{h}_{t-1}) \tag{7}$$

$$\overleftarrow{h}_t = GRU(f_t, \overleftarrow{h}_{t+1}) \tag{8}$$

The final semantic vector $\vec{u}$ is typically obtained by concatenating the last hidden state of the forward GRU and the first hidden state of the backward GRU, as shown in Equation (9).

$$\vec{u} = [\vec{h}_{|F|}; \overleftarrow{h}_1] \tag{9}$$

The semantic vector , obtained through this process, encapsulates the information of all factual elements derived from the case description. This vector will subsequently be integrated into the inference model to enhance the model's prediction capabilities regarding legal charges.

**4.2 Knowledge matching**

This section introduces the second module of our method. This module matches case description $X$ with a given knowledge base $\mathcal{K}$. The knowledge snippets matched serve as prompts to enhance the reasoning model's prediction of legal charges.

We use *THUOCL_Law* as the knowledge base. *THUOCL_Law* is a subbase of the Tsinghua University Open Chinese Lexicon (THUOCL), a high-quality Chinese lexicon compiled and launched by the Natural Language Processing and Social Humanities Computing Laboratory of Tsinghua University, in which all subbases have undergone multiple rounds of manual screening to ensure the accuracy. Table 1 shows some of the knowledge in *THUOCL_Law*.

**Table 1** Part of the knowledge in THUOCL_Law

| |
|---|
| 违背妇女意志(against women's will)，违约(breach of contract)，拐卖(kidnapping)，抢夺(snatch)，殴打(beat up)，致残(disabled)，故意(deliberately)，残忍(cruel) |

As presented in the table, a knowledge snippet is essentially a keyword. As [35] discussed, keywords in case descriptions are crucial for predictions of legal charges. We simply utilize regular expressions to match these keywords from the case description $X$, thereby obtaining a list of keywords $K = [k_1, k_2, ..., k_{|K|}]$, also referred to as the list of knowledge snippets.

The concatenation of the knowledge snippets in $K$ will serve as a prompt, and in conjunction with other components, act as the input for the inference model.

## 4.3 Legal language model reasoning

This section introduces the third module of our method. In this module, we employ a legal language model to reason the legal charge associated with the given case description $X$. Traditionally, the task of predicting legal charges is viewed as a classification problem, where the model's output is directly a probability distribution, and the index of the highest probability is the predicted label. However, we transform the task of legal charge prediction into a language modeling (cloze test) task, prompting the model to predict masked tokens. Then, we map the predictions at these masked positions onto the final category labels.

To implement the language modeling task, we construct hard prompt templates $T_1$ and $T_2$, as follows:

$T_1$ = "*He will be charged with criminal responsibility for*"

$T_2$ = "*Keywords in the case description are as follows:*"

These hard prompts serve as part of the input for the inference model, guiding the model to predict masked tokens. In addition to hard prompts, we also incorporate two soft prompts $s_1$ and $s_2$ into the input. Semantic vector $\vec{u}$ obtained in Section 4.1 will be merged with these soft prompts, injecting the knowledge about factual elements into the model's inference. Moreover, the masked tokens $M = [m_1, m_2, ..., m_{|M|}]$ and case description $X = [x_1, x_2, ..., x_L]$ are also essential components of the input. Lastly, the concatenation of knowledge snippets $K$, acquired in Section 4.2, is also included as a part of the input.

In summary, the input for the inference model is composed of the case description $X$, hard prompt texts $T_1$ and $T_2$, soft prompts $s_1$ and $s_2$, the masked sequence $M = [m_1, m_2, ..., m_{|M|}]$, and the concatenation of knowledge snippets $K$, as illustrated in Equation (10).

$$X' = [s_1, t_{1,1}, t_{1,2}, ..., t_{1,|T_1|}, m_1, m_2, ..., m_{|M|}, x_1, x_2, ... x_L, t_{2,1}, t_{2,2}, ..., t_{2,|T_2|}, k_1, k_2, ..., k_{|K|}, s_2] \quad (10)$$

Where, $t_{1,i}$ represents the $i_{th}$ token in $T_1$, and $t_{2,i}$ denotes the $i_{th}$ token in $T_2$. Besides, $m_i$ stands for the $i_{th}$ mask, $x_i$ refers to the $i_{th}$ token in $X$, and $k_i$ indicates the $i_{th}$ token in $K$.

To facilitate understanding, we illustrate $X'$ more intuitively through the example provided in Figure 3. In the diagram, two purple tokens represent the soft prompts, the green section is $T_1$ and the blue section is $T_2$. The red tokens indicate the masked tokens, the black text is the original case description $X$, and the yellow section is the concatenation of keywords.

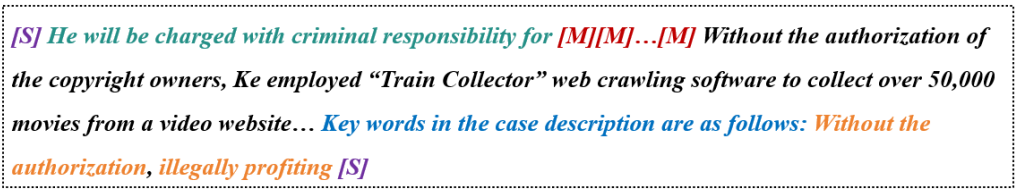

**Fig. 3** Schematic diagram of the components of $X'$

Nextly, we input $X'$ into the inference model, which is a pre-trained legal language model. The inference model consists of embedding and encoding layers, as shown in Figure 1. During the embedding layer stage, all tokens in $X'$ except the soft prompts are embedded by the embedding layer of the inference model. At the same time, the soft prompts in $X'$ are embedded by an additional trainable embedding matrix. This process is shown in Equation (11).

$$\vec{e}_i = \begin{cases} S[i], & if\ i \in soft_{idx} \\ Embedding(token_i), & otherwise \end{cases} \quad (11)$$

Where $S \in \mathbb{R}^{|soft_{idx}| \times d_h}$ is a trainable embedding matrix, and $soft_{idx}$ is the index of the soft prompt token. $d_h$ is embedding dimension of the model. Therefore, an embedding vector sequence $E$ of all the tokens (including soft prompt tokens) in $X'$ can be obtained by the equation.

$$E = [\vec{e}_{s_1}, \vec{e}_{t_{1,1}}, \vec{e}_{t_{1,2}}, ..., \vec{e}_{t_{1,|T_1|}}, \vec{e}_{m_1}, \vec{e}_{m_2}, ..., \vec{e}_{m_{|M|}}, \vec{e}_{x_1}, \vec{e}_{x_2}, ..., \vec{e}_{x_L}, \vec{e}_{t_{2,1}}, ..., \vec{e}_{t_{2,|T_2|}}, \vec{e}_{k_1}, \vec{e}_{k_2}, ..., \vec{e}_{k_{|K|}}, \vec{e}_{s_2}]$$

Where $\vec{e}_{t_{1,i}}$, $\vec{e}_{t_{2,i}}$, $\vec{e}_{m_i}$, $\vec{e}_{x_i}$ and $\vec{e}_{k_i}$ denote the embedding vectors of the $i_{th}$ tokens in $T_1$, $T_2$, $M$, $X$, and $K$ respectively. Besides, $\vec{e}_{s_1}$ and $\vec{e}_{s_2}$ denote the embedding vectors of the soft prompt tokens $s_1$ and $s_2$.

To inject the inference model with factual element information during its forward computation, we add the semantic vector $\vec{u}$ obtained in Section 4.1 to the vectors $\vec{e}_{s_1}$ and $\vec{e}_{s_2}$ respectively. This results in prompt vectors enriched with factual element information. This process is demonstrated in Equations (12) and (13).

$$\vec{e}'_{s_1} = \vec{e}_{s_1} + \vec{u} \quad (12)$$

$$\vec{e}'_{s_2} = \vec{e}_{s_2} + \vec{u} \quad (13)$$

Subsequently, we replace $\vec{e}_{s_1}$ and $\vec{e}_{s_2}$ in $E$ using $\vec{e}'_{s_1}$ and $\vec{e}'_{s_2}$ to obtain:

$$E' = [\vec{e}'_{s_1}, \vec{e}_{t_{1,1}}, \vec{e}_{t_{1,2}}, ..., \vec{e}_{t_{1,|T_1|}}, \vec{e}_{m_1}, \vec{e}_{m_2}, ..., \vec{e}_{m_{|M|}}, \vec{e}_{x_1}, \vec{e}_{x_2}, ..., \vec{e}_{x_L}, \vec{e}_{t_{2,1}}, ..., \vec{e}_{t_{2,|T_2|}}, \vec{e}_{k_1}, \vec{e}_{k_2}, ..., \vec{e}_{k_{|K|}}, \vec{e}'_{s_2}]$$

Finally, we input $E'$ into the encoding layer of the inference model and obtain the hidden layer outputs of the model, that is, the contextual representations for each token. This process is shown as Equation (14).

$$R = \vec{r}_{s_1}, \vec{r}_{t_{1,1}}, \vec{r}_{t_{1,2}}, ..., \vec{r}_{t_{1,|T_1|}}, \vec{r}_{m_1}, \vec{r}_{m_2}, ..., \vec{r}_{m_{|M|}}, \vec{r}_{x_1}, \vec{r}_{x_2}, ..., \vec{r}_{x_L}, \vec{r}_{t_{2,1}}, ..., \vec{r}_{t_{2,|T_2|}}, \vec{r}_{k_1}, \vec{r}_{k_2}, ..., \vec{r}_{k_{|K|}}, \vec{r}_{s_2} = Encoding(E') \quad (14)$$

The model's objective is to predict the tokens at the masked position. Therefore, by projecting $\vec{r}_{m_1}, \vec{r}_{m_2}, ..., \vec{r}_{m_{|M|}}$ into the vocabulary space, we can obtain the probability distributions of the predicted tokens. Subsequently, by selecting the indices with the highest probabilities, the model can determine the tokens at the masked positions. We denote the $i_{th}$ predicted token at the masked positions as $\hat{m}_i$.

The next section describe the process of mapping predicted tokens $\hat{m}_1, \hat{m}_2, ..., \hat{m}_{|M|}$ to a charge category label.

**4.4 Legal charge category mapping**

**Table 2** Some of the label texts

| |
|---|
| 制造、贩卖、传播淫秽物品 (Manufacture, sell, and disseminate obscene materials) |
| 非法持有、私藏枪支、弹药 (Illegal possession of firearms and ammunition) |
| 非法占用农用地 (Illegal occupation of agricultural land) |
| 非法种植毒品原植物 (Illegal cultivation of narcotic plants) |
| 危害公共安全 (Endanger public safety) |

We need to map the outcomes of the language modeling (cloze test) task back to the classification task. To this end, this section constructs a mapping from predicted tokens to legal charge categories. We calculate the Jaccard similarities between the predicted tokens and the texts of the legal charge labels. For instance, if the predicted tokens has the highest Jaccard similarity with "Manufacture, sell, and disseminate obscene materials", then the legal charge predicted by the model for the current case description is "Manufacture, sell, and disseminate obscene materials". Table 2 shows some of the texts of the legal charge labels.

Therefore, the final prediction of the model is shown in Equation (15).

$$\hat{y} = argmax(Jaccard(\hat{m}_{1:|M|}, v_y)) \tag{15}$$

Where $\hat{y}$ is the final label predicted by the model, and $\hat{m}_{1:|M|}$ denotes the predicted tokens at the masked positions, and $v_y$ denotes the text of category $y$.

## 5. Experimental settings

This section details the experimental setup, including the datasets used, baselines, and implementation specifics of our method.

### 5.1 Datasets

We utilized CAIL-2018 [7], the largest Chinese legal charge prediction dataset, as our experimental dataset. The authors acquired 2.6 million public criminal cases from the Supreme People's Courts of China and subjected them to preprocessing, ultimately obtaining 2,676,075 case description texts accompanied by 196 unique legal charge labels. Each case description corresponds to only one legal charge label, so the task is a single-label classification problem. Table 3 shows some instances from the CAIL2018 dataset. In addition, 2/3 of the total are used as the training set and the remaining 1/3 as the test set.

**Table 3** Some instances in the CAIL2018 dataset.

| Case description | Legal charge label |
| --- | --- |
| 被告人罗某甲…，罗某甲踢了项某乙一脚，之后双方发生互殴，…<br>Defendant Luo…, Luo kicked Xiang, and then the two sides fought each other, ... | 故意伤害<br>Intentional injury |
| 被告人黄某携带作案工具螺丝…，后转售后得赃款…<br>Defendant Huang carried a screwdriver as a crime tool..., and then resold it for money... | 盗窃<br>Theft |
| 被告人周某在本县武康街道营盘小区…窃得黑色苹果 7PLUS 手机一部…<br>Defendant Zhou stole a black Apple 7PLUS mobile phone from Yingpan Community, Wukang Street... | 盗窃<br>Theft |

### 5.2 Baselines

We compare the proposed method against the following baselines to verify its advancement.

(1) **CNN** [1]: The method proposed by Wang et al., which uses convolutional neural networks to model the terminologies and keywords within case descriptions.

(2) **BiLSTM** [2]: The method proposed by Yang et al., which models the global semantics of case descriptions via bidirectional long and short-term memory networks.

(3) **BiGRU** [3]: The method proposed by Chen et al. Compared to bidirectional long and short-term memory networks, bidirectional gated networks have lower computational complexity for modeling the global semantics of case descriptions.

(4) **Attention** [5]: The attention mechanism-based method proposed by Sukanya and Priyadarshini. Attention mechanisms can assign different weights to different factual information in a case description. This method achieved state-of-the-art results across multiple benchmarks.

(5) **BERT** [35]: The method of fine-tuning a BERT on the legal charge prediction task. This is a powerful baseline.

(6) **HMN** [6]: This is a hierarchical matching network for crime classification proposed by Wang et al. This method is a novel and strong baseline on the CAIL2018 dataset.

(7) **ChatGLM** [39]: We directly engage in dialogue with the Chinese conversational LLM, ChatGLM, to obtain the legal charge label corresponding to each case description.

### 5.3 Implementation specifics

All our experiments were conducted on a 40G A100 GPU. During the training phase, the model employed a learning rate of 1e-5, a batch size of 8. We employ Lawformer [22], a recently proposed legal pre-trained language model, as our inference model, which can accept text inputs up to a maximum length of 4,000. The number of masked tokens to be predicted is set to 20. Besides, we set a maximum training epoch of 50 with an early stopping mechanism.

## 6. Results and discussion

This section discusses the experimental results. Section 6.1 compares our method with baselines, while Section 6.2 validates the effectiveness of each module within our method through ablation experiments. Moreover, Section 6.3 analyzes the impact of the training data size on the performance of our method. Section 6.4 analyzes the hyperparameter settings, and Section 6.5 validates the interpretability of our model through a specific case study.

### 6.1 Comparison with baselines

Table 4 Performance of the baselines and our method

| Method  | P    | R    | F1   |
|---------|------|------|------|
| CNN     | 0.72 | 0.70 | 0.71 |
| BiLSTM  | 0.69 | 0.67 | 0.68 |
| BiGRU   | 0.66 | 0.66 | 0.66 |
| Attention | 0.77 | 0.76 | 0.76 |
| BERT    | 0.62 | 0.61 | 0.61 |
| HMN     | 0.79 | 0.77 | 0.78 |
| ChatGLM | 0.69 | 0.68 | 0.68 |
| **Ours** | **0.85** | **0.83** | **0.84** |

Table 4 displays the performance of our method compared to the baselines. From the table, it can be observed that the performances of **CNN**, **BiLSTM**, and **BiGRU** are relatively similar, with macro F1 scores ranging between 0.66 and 0.71. **CNN** performs the best, which may be attributed to the fact that case descriptions are often lengthy, and thus modeling the sequence structure is less effective than modeling key information. Among the traditional neural network models presented in the first four rows, **Attention** achieves the best performance. Attention mechanisms are capable of focusing on local key information within case descriptions as well as the correlations between different pieces of information, hence achieving performance significantly beyond that of other traditional neural network baselines. Based on this, we can also infer that factual elements and knowledge snippets within case descriptions can enhance the effectiveness of legal charge prediction.

From the fifth row of the table, it is evident that the usually strong baseline **BERT** performs worse than traditional neural network models, achieving only a macro F1 score of 0.61. This is due to BERT's input length limitation of 512, which prevents it from fully modeling case description texts that average over 1,000 in length. In contrast, traditional neural network models do not have an input length restriction, enabling them to model case descriptions more completely and thereby achieve better results than **BERT**. Our method employs Lawformer as the inference model, which can accept case description inputs of over 4000 in length, thereby enabling more complete modeling of case descriptions.

From the sixth row of the table, it is apparent that the performance of **HMN** surpasses other baselines due to its integration of hierarchical matching, which not only allows for the complete modeling of the entire case description but also enables hierarchical modeling of key information. Furthermore, from the seventh row of the table, we can observe that the performance of **ChatGLM** is mediocre. Despite ChatGLM having learned a vast amount of general semantic knowledge and possessing good zero-shot reasoning capabilities, it still exhibits hallucination issues in specialized fields such as law and medicine.

Finally, from the last row of the table, we can see that our method achieved the best results, with a macro F1 score reaching 0.84. This demonstrates that our method is feasible and effective. Furthermore, it suggests that factual elements obtained via relevant articles and the conversational LLM, as well as knowledge snippets acquired from the knowledge base, have enhanced the model's ability to predict legal charges. To further validate the contribution of each module in our method, we conducted ablation experiments in Section 4.2.

## 6.2 Ablation experiments

This section analyzes the contribution of each module in our method through ablation experiments, with the experimental results shown in Table 5. The first row of the table represents the performance of our original method on the CAIL-2018 dataset.

**Table 5** Results of the ablation experiments

|  | P | R | F1 |
|---|---|---|---|
| Ours | 0.85 | 0.83 | 0.84 |
| − knowledge snippets | 0.82 | 0.81 | 0.82 (-0.02) |
| − factual elements | 0.81 | 0.81 | 0.81 (-0.03) |
| − knowledge snippets and factual elements | 0.79 | 0.77 | 0.78 (-0.06) |
| − Contrastive training | 0.83 | 0.82 | 0.82 (-0.02) |

Firstly, we removed the knowledge snippets module, with the experimental results shown in the second row of the table. It can be observed that the macro F1 score of the model decreased by 0.02, indicating that focusing on factual elements within case descriptions indeed enhances the prediction effectiveness of legal charges.

Nextly, we retained the knowledge snippets module and removed the factual element acquisition and encoding module. It can be seen from the third row of the table that this resulted in a decrease of 0.03 in the model's macro F1 score. This indicates that factual elements within case descriptions also contribute beneficially to the legal charge prediction task.

Finally, we removed both modules simultaneously, with the experimental results shown in the fourth row of the table. It can be observed that the performance of the model significantly decreased by 0.06.

We also validated the role of contrastive training, as discussed in Section 4.1, in accurately retrieving relevant legal articles. We removed the learning of a joint semantic space and directly used RoBERTa to encode the case descriptions and legal articles. We still determined the relevance between a given case description and each legal article by calculating the inner product between their encoded vectors. The experimental results indicate that this operation led to a decrease of 0.02 in the final results of the model, as shown in the last row of the table. This demonstrates that it is necessary to train a joint semantic space for case descriptions and legal articles through contrastive learning.

## 6.3 Impact of training data volume

This section tests the effects of using training data of different scales on the model and analyzes the experimental results. Figure 4 illustrates the variation curves of the models' F1 scores on the test set as the training data decreases. It can be observed that with the reduction of the training data volume, the macro F1 scores of all methods decline (except for **ChatGLM**, as it does not require training data), yet the decrease in the macro F1 score of our method is more gradual. This indicates that our approach possesses a stronger advantage in scenarios of data scarcity. Furthermore, it was found that even when the training data volume was reduced to merely 10% of its original size, our method still achieved a macro F1 score that surpassed that of ChatGLM. This can be attributed to two factors. On one hand, the CAIL-2018 dataset itself is quite large, meaning that even 10% of the original training data volume consists of 80,000 training samples. On the other hand, this suggests that conversational LLMs still do not hold a definitive advantage in specific domains such as law and medicine.

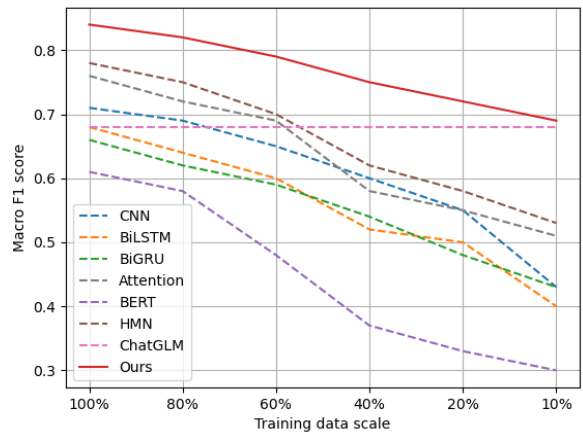

Fig. 4 Variation curves of the models' F1 scores on the test set as the training data decreases.

## 6.4 Hyperparameter Analysis

This section analyzes the settings of hyperparameters in our method, with the experimental results shown in Figure 5. The subgraphs from left to right correspond to the number of relevant articles searched being 2, 4, 6, and 8, respectively. The vertical axis of each subgraph represents macro F1 score. Overall, it can be observed that the model achieves the best F1 score when the number of relevant articles searched is 4. This indicates that searching too many relevant articles may introduce noise to the conversational LLM's process of extracting factual elements from case descriptions.

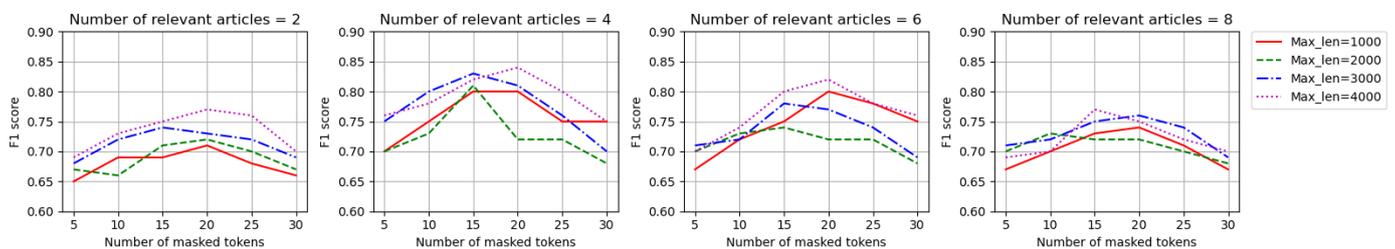

Fig. 5   Influence of hyperparameters on our model

Additionally, we examine each subgraph. The different colored lines within each subgraph represent the maximum truncation length of case descriptions, set at 1000, 2000, 3000, and 4000, respectively. It is evident that as the maximum truncation length of case descriptions increases, the model's performance improves. This is because the larger the maximum truncation length, the more complete the model's semantic understanding of case descriptions, thereby achieving better results. Lastly, the horizontal axis in each subgraph represents the number of masked tokens to be predicted. It can be observed from each subgraph that the model tends to achieve higher F1 values when the number of masked tokens is set to 15 and 20. Therefore, we set the number of masked tokens to be predicted by the model to 20.

**6.5 Case study**

In this section, we qualitatively analyze the prediction of our method for a given case description to validate the interpretability and effectiveness of the model. In Figure 6, a case description is provided with its true label identified as "Rape." From the figure, it is evident that our method's knowledge matching module successfully matched the following knowledge snippets from the case description: 发生性关系(occurrence of sexual relation), 暴力(violence), and 威胁(threat). Intuitively, we can agree that these three keywords are highly related to the crime of rape. Therefore, it can be said that the knowledge matching module provides evidence for the model's prediction. This demonstrates that our proposed method has a certain degree of interpretability.

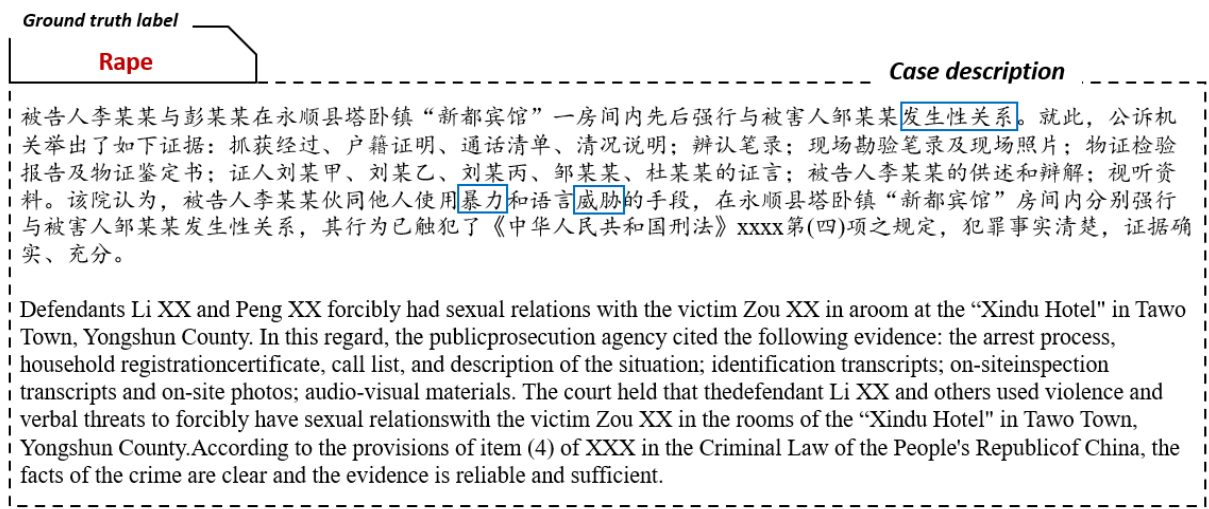

**Fig. 6** Analysis of the prediction process of our method on a given case description.

Additionally, the factual element acquisition module successfully extracted the factual elements within the case description via the conversational LLM. These factual elements include:

- **Location**: In a room at "New Capital Hotel" in Tawo Town, Yongshun County.
- **Sexual acts**: Defendant Li and Defendant Peng forcibly engaged in sexual acts with the victim Zou in the room at "New Capital Hotel."
- **Means used to commit the crime**: Defendant Li, along with others, used violence and verbal threats.
- **Crime time**: The specific time is not mentioned.

These factual elements also contribute to providing interpretability for legal professionals or users without legal expertise. In summary, this section demonstrates that our method exhibits a high level of interpretability, a crucial aspect in the legal domain.

# 7. Conclusion

We propose a multi-source heterogeneous knowledge-enhanced prompt learning method for legal charge prediction. We transform the legal charge prediction task from a classification problem into a masked language modeling problem, employing prompt learning for model training. Subsequently, the method extracts knowledge snippets from case descriptions via a legal knowledge base and obtains factual elements through relevant legal articles and a conversational LLM. By injecting factual elements and knowledge snippets into the model's reasoning in different ways, the model's understanding of case descriptions is enhanced. We also introduce a joint semantic space learning for case descriptions and legal articles using a contrastive loss to more accurately identify legal articles relevant to a case description. Experimental results demonstrate that our approach achieves the best performance and exhibits low training data dependency. Despite the popularity of conversational LLMs, our method outperforms them in this task. Additionally, experiments show that our method maintains good interpretability, a crucial aspect in legal charge prediction tasks.

This study does not explore situations where a single case description may correspond to multiple legal charges. Future research will delve into predictive tasks involving multiple legal charges.

**Code availability**

The code in this paper is available by contacting the corresponding authors.

**CRediT authorship contribution statement**

Jingyun Sun: Conceptualization, Methodology, Formal analysis, Investigation, Writing – original draft, Writing – review & editing, Resources. Chi Wei: Supervision, Investigation, Writing – review & editing.

**Declaration of Competing Interest**

The authors have no relevant financial or non-financial interests to disclose.

**Data availability**

Data will be made available on request.

# Reference

[1] H. Wang, T. He, Z. Zou, S. Shen, and Y. Li, "Using case facts to predict accusation based on deep learning," in *2019 IEEE 19th International Conference on Software Quality, Reliability and Security Companion (QRS-C)*, IEEE, 2019, pp. 133–137.

[2] Z. Yang, P. Wang, L. Zhang, L. Shou, and W. Xu, "A recurrent attention network for judgment prediction," in *Artificial Neural Networks and Machine Learning–ICANN 2019: Text and Time Series: 28th International Conference on Artificial Neural Networks, Munich, Germany, September 17–19, 2019, Proceedings, Part IV 28*, Springer, 2019, pp. 253–266.

[3] H. Chen, D. Cai, W. Dai, Z. Dai, and Y. Ding, "Charge-Based Prison Term Prediction with Deep Gating Network," in *Proceedings of the 2019 Conference on Empirical Methods in Natural Language Processing and the 9th International Joint Conference on Natural Language Processing (EMNLP-IJCNLP)*, 2019, pp. 6362–6367.

[4] I. Chalkidis, I. Androutsopoulos, and N. Aletras, "Neural Legal Judgment Prediction in English," in *Proceedings of the 57th Annual Meeting of the Association for Computational Linguistics*, 2019, pp. 4317–4323.

[5] G. Sukanya and J. Priyadarshini, "A meta analysis of attention models on legal judgment prediction system," *International Journal of Advanced Computer Science and Applications*, vol. 12, no. 2, 2021.

[6] P. Wang, Y. Fan, S. Niu, Z. Yang, Y. Zhang, and J. Guo, "Hierarchical matching network for crime classification," in *proceedings of the 42nd international ACM SIGIR conference on research and development in information retrieval*, 2019, pp. 325–334.

[7] C. Xiao *et al.*, "Cail2018: A large-scale legal dataset for judgment prediction," *arXiv preprint arXiv:1807.02478*, 2018.

[8] R. Keown, "Mathematical models for legal prediction," *Computer/lj*, vol. 2, p. 829, 1980.

[9] F. Kort, "Predicting Supreme Court decisions mathematically: A quantitative analysis of the 'right to counsel' cases," *American Political Science Review*, vol. 51, no. 1, pp. 1–12, 1957.

[10] S. S. Nagel, "Applying correlation analysis to case prediction," *Tex. L. Rev.*, vol. 42, p. 1006, 1963.

[11] T. Mikolov, I. Sutskever, K. Chen, G. S. Corrado, and J. Dean, "Distributed representations of words and phrases and their compositionality," *Advances in neural information processing systems*, vol. 26, 2013.

[12] Z. Hu, X. Li, C. Tu, Z. Liu, and M. Sun, "Few-shot charge prediction with discriminative legal attributes," in *Proceedings of the 27th International Conference on Computational Linguistics*, 2018, pp. 487–498.

[13] I. Chalkidis and D. Kampas, "Deep learning in law: early adaptation and legal word embeddings trained on large corpora," *Artificial Intelligence and Law*, vol. 27, no. 2, pp. 171–198, 2019.

[14] I. Chalkidis, M. Fergadiotis, P. Malakasiotis, and I. Androutsopoulos, "Large-scale multi-label text classification on EU legislation," *arXiv preprint arXiv:1906.02192*, 2019.

[15] C.-L. Liu, C.-T. Chang, and J.-H. Ho, "Case instance generation and refinement for case-based criminal summary judgments in Chinese.," *Journal of Information Science and Engineering*, vol. 20, no. 4, pp. 783–800, 2004.

[16] P. Bhattacharya, K. Ghosh, A. Pal, and S. Ghosh, "Legal case document similarity: You need both network and text," *Information Processing & Management*, vol. 59, no. 6, p. 103069, 2022.

[17] S. Bi, Z. Ali, M. Wang, T. Wu, and G. Qi, "Learning heterogeneous graph embedding for Chinese legal document similarity," *Knowledge-Based Systems*, vol. 250, p. 109046, 2022.

[18] L. Gan, K. Kuang, Y. Yang, and F. Wu, "Judgment prediction via injecting legal knowledge into neural networks," in *Proceedings of the AAAI Conference on Artificial Intelligence*, 2021, pp. 12866–12874.

[19] Y. Chen, Y. Sun, Z. Yang, and H. Lin, "Joint entity and relation extraction for legal documents with legal feature enhancement," in *Proceedings of the 28th International Conference on Computational Linguistics*, 2020, pp. 1561–1571.

[20] P. Clark, O. Tafjord, and K. Richardson, "Transformers as soft reasoners over language," in *Proceedings of the Twenty-Ninth International Conference on International Joint Conferences on Artificial Intelligence*, 2021, pp. 3882–3890.

[21] N. Xu, P. Wang, L. Chen, L. Pan, X. Wang, and J. Zhao, "Distinguish Confusing Law Articles for Legal Judgment Prediction," in *Proceedings of the 58th Annual Meeting of the Association for Computational Linguistics*, 2020, pp. 3086–3095.

[22] C. Xiao, X. Hu, Z. Liu, C. Tu, and M. Sun, "Lawformer: A pre-trained language model for chinese legal long documents," *AI Open*, vol. 2, pp. 79–84, 2021.

[23] T. Schick and H. Schütze, "Exploiting Cloze-Questions for Few-Shot Text Classification and Natural Language Inference," in *Proceedings of the 16th Conference of the European Chapter of the Association for Computational Linguistics: Main Volume*, 2021, pp. 255–269.

[24] H. Wu, B. Ma, W. Liu, T. Chen, and D. Nie, "Fast and constrained absent keyphrase generation by prompt-based learning," in *Proceedings of the AAAI Conference on Artificial Intelligence*, 2022, pp. 11495–11503.

[25] T. Shin, Y. Razeghi, R. L. Logan IV, E. Wallace, and S. Singh, "AutoPrompt: Eliciting Knowledge from Language Models with Automatically Generated Prompts," in *Proceedings of the 2020 Conference on Empirical Methods in Natural Language Processing (EMNLP)*, 2020, pp. 4222–4235.

[26] R. Song *et al.*, "Label prompt for multi-label text classification," *Applied Intelligence*, vol. 53, no. 8, pp. 8761–8775, 2023.

[27] X. Chen *et al.*, "Relation Extraction as Open-book Examination: Retrieval-enhanced Prompt Tuning," in *Proceedings of the 45th International ACM SIGIR Conference on Research and Development in Information Retrieval*, 2022, pp. 2443–2448.

[28] X. Han, W. Zhao, N. Ding, Z. Liu, and M. Sun, "Ptr: Prompt tuning with rules for text classification," *AI Open*, vol. 3, pp. 182–192, 2022.

[29] G. Jiang, S. Liu, Y. Zhao, Y. Sun, and M. Zhang, "Fake news detection via knowledgeable prompt learning," *Information Processing & Management*, vol. 59, no. 5, p. 103029, 2022.

[30] J. D. M.-W. C. Kenton and L. K. Toutanova, "BERT: Pre-training of Deep Bidirectional Transformers for Language Understanding," in *Proceedings of NAACL-HLT*, 2019, pp. 4171–4186.

[31] Y. Liu *et al.*, "Roberta: A robustly optimized bert pretraining approach," *arXiv preprint arXiv:1907.11692*, 2019.

[32] H. Yang, X.-Y. Liu, and C. D. Wang, "FinGPT: Open-Source Financial Large Language Models," *arXiv preprint arXiv:2306.06031*, 2023.

[33] K. Jiang, R. Pradeep, and J. Lin, "Exploring listwise evidence reasoning with t5 for fact verification," in *Proceedings of the 59th Annual Meeting of the Association for Computational Linguistics and the 11th International Joint Conference*

*on Natural Language Processing (Volume 2: Short Papers)*, 2021, pp. 402–410.

[34] T. Zhu, Y. Qin, Q. Chen, B. Hu, and Y. Xiang, "Enhancing entity representations with prompt learning for biomedical entity linking," in *Proceedings of the Thirty-First International Joint Conference on Artificial Intelligence*, 2021, pp. 4036–4042.

[35] H. Zhong, C. Xiao, C. Tu, T. Zhang, Z. Liu, and M. Sun, "How Does NLP Benefit Legal System: A Summary of Legal Artificial Intelligence," in *Proceedings of the 58th Annual Meeting of the Association for Computational Linguistics*, 2020, pp. 5218–5230.

[36] F. Giunchiglia, M. Yatskevich, and P. Shvaiko, "Semantic matching: Algorithms and implementation," in *Journal on data semantics IX*, Springer, 2007, pp. 1–38.

[37] L. Otero-Cerdeira, F. J. Rodríguez-Martínez, and A. Gómez-Rodríguez, "Ontology matching: A literature review," *Expert Systems with Applications*, vol. 42, no. 2, pp. 949–971, 2015.

[38] T. Wang and P. Isola, "Understanding contrastive representation learning through alignment and uniformity on the hypersphere," in *International Conference on Machine Learning*, PMLR, 2020, pp. 9929–9939.

[39] J. Cui, Z. Li, Y. Yan, B. Chen, and L. Yuan, "Chatlaw: Open-source legal large language model with integrated external knowledge bases," *arXiv preprint arXiv:2306.16092*, 2023.